\documentclass[preprint,12pt]{elsarticle}

\usepackage{amssymb}
\usepackage{amsmath}

\usepackage[printonlyused]{acronym}
\usepackage[hang,flushmargin]{footmisc}
\newcommand\blfootnote[1]{%
  \begingroup
  \renewcommand\thefootnote{}\footnote{#1}%
  \addtocounter{footnote}{-1}%
  \endgroup
}
\usepackage{subfig}

\journal{Energy \& AI}

\begin{document}

\begin{frontmatter}



\title{Transfer learning applications for anomaly detection in wind turbines}


\author[address1]{Cyriana M.A. Roelofs\corref{cor1}}
\ead{cyriana.roelofs@iee.fraunhofer.de}
\author[address1]{Christian G{\"u}ck}
\ead{christian.gueck@iee.fraunhofer.de}
\author[address1]{Stefan Faulstich}
\ead{stefan.faulstich@iee.fraunhofer.de}
\cortext[cor1]{Corresponding author}

\address[address1]{Fraunhofer IEE, Joseph-Beuys-Straße 8, 34117 Kassel, Germany}

\begin{abstract}

Anomaly detection in wind turbines typically involves using normal behaviour models to detect faults early. However, training autoencoder models for each turbine is time-consuming and resource intensive. Thus, transfer learning becomes essential for wind turbines with limited data or applications with limited computational resources. This study examines how cross-turbine transfer learning can be applied to autoencoder-based anomaly detection. 
Here, autoencoders are combined with constant thresholds for the reconstruction error to determine if input data contains an anomaly. The models are initially trained on one year's worth of data from one or more source wind turbines. They are then fine-tuned using smaller amounts of data from another turbine. Three methods for fine-tuning are investigated: adjusting the entire autoencoder, only the decoder, or only the threshold of the model. The performance of the transfer learning models is compared to baseline models that were trained on one year's worth of data from the target wind turbine.
The results of the tests conducted in this study indicate that models trained on data of multiple wind turbines do not improve the anomaly detection capability compared to models trained on data of one source wind turbine.
In addition, modifying the model's threshold can lead to comparable or even superior performance compared to the baseline, whereas fine-tuning the decoder or autoencoder further enhances the models' performance. 

\end{abstract}




\end{frontmatter}


\section{Introduction}\label{introduction}

\blfootnote{Nomenclature:
\begin{acronym}
\acro{tl}[TL]{Transfer Learning}
\acro{ae}[AE]{autoencoder}
\acro{ann}[ANN]{artificial neural network}
\acro{arcana}[ARCANA]{Anomaly Root Cause Analysis}
\acro{ml}[ML]{machine learning}
\acro{mse}[MSE]{mean square error}
\acro{rmse}[RMSE]{root mean square error}
\acro{nbm}[NBM]{normal behaviour model}
\acro{om}[O\&M]{Operation and maintenance}
\acro{opmode}[OP-mode]{operational mode}
\acro{re}[RE]{AE reconstruction error}
\acro{scada}[SCADA]{Supervisory Control and Data Acquisition}
\end{acronym}
}

Wind power plays a crucial role in addressing climate change, offering significant promise among renewable energy sources. However, wind farms are often situated in regions with challenging weather conditions, leading to complex operating conditions and potential failures in wind turbines. This results in increased maintenance costs and decreased efficiency. To mitigate these challenges, there has been a growing focus on condition monitoring and early fault detection using machine learning models for wind turbines \cite{Stetco.2019}. Numerous approaches concentrate on training a \ac{nbm} using wind turbine \ac{scada} data, to identify anomalous behaviour or detect specific failures in wind turbines.

Training an anomaly detection model based on NBM requires enough data representing normal operation of the wind turbine. However, such data is not available for new installations or after maintenance actions, such as an oil refill or component replacements resulting in changed normal behaviour. To overcome the scarcity of training data, the application of \ac{tl} can be beneficial.
\ac{tl} involves transferring learned behaviour from one model to another, similar to how humans learn new tasks by applying knowledge from related tasks. In the context of wind turbines, this involves using a pretrained model for anomaly detection in one wind turbine (or a group of wind turbines) and repurposing it for anomaly detection in another wind turbine. This approach saves significant time and resources that would otherwise be required for training a new model \cite{Weiss.2016, Zhuang.2021}. 

In this study, our focus is on exploring the applications of \ac{tl} for autoencoder-based anomaly detection in wind turbine data. We first explore \ac{tl} within one wind farm, to test whether anomaly detection is still possible for an wind turbine for which only a small amount of training data is available. Specifically, a model is trained on data of one wind turbine and transferred to another wind turbine within the same wind farm. We examine the impact of different amounts of tuning data for different wind turbine types, using various \ac{tl} methods.
Additionally, we use data of multiple wind turbines to train a \ac{nbm} and fine-tune it on data of a single target wind turbine which is not in the training dataset. The hypothesis is that this model can generalise better and be better tuned to the target task. In both cases, we validate the transferred models' ability to detect anomalies by evaluating three case studies, demonstrating that early fault detection in wind turbines is indeed possible even with a limited amount of training data.

The paper is structured as follows. First we present an overview of related work in section \ref{related_work}. In section \ref{method} we describe the data and models used and the \ac{tl} methods which we applied. Next, the results are presented and discussed in \ref{results}. The results of the transferred models are first compared to their corresponding baseline models, to show they are capable of distinguishing normal from anomalous behaviour almost or as well as the baseline model. Afterwards, we demonstrate the model ability to detect failures early using three case studies. Finally, the paper is summarised in section \ref{summary}.

\section{Related work}\label{related_work}


Research on \ac{tl} applied to unsupervised or semi-supervised (only using ‘healthy’ data) machine learning, such as anomaly detection, is limited \cite{Michau.2021,Zgraggen.2021}. Most work focuses on dimensionality reduction and clustering, which is often not applicable to anomaly detection in predictive maintenance settings. Due to limited occurrences of failures, anomaly detection models are often hard to train. Here, Transfer Learning approaches may help overcome this problem \cite{Chen.2019}. In addition, the training of wind turbine-specific models is not always feasible. For example, if a wind turbine is newly installed or not all historical data is informative due to re-calibration or component replacement(s). Also, training ML models on only one turbine greatly reduces the amount of training time for the whole wind farm (containing up to hundreds of turbines) \cite{Zgraggen.2021}.

Zgraggen et al. present three \ac{tl} approaches for anomaly detection in wind turbines using deep learning methods \cite{Zgraggen.2021}. The anomaly detection model used is a normal behaviour model (NBM) of a regression CNN that has 4 SCADA parameters as input and predicts the gearbox bearing temperature. Three \ac{tl} methods are tested for two scenarios: cross-turbine \ac{tl} within the same wind farm and cross-turbine \ac{tl} across different wind farms. For the first scenario the simplest method, where an additional linear regression (LR) is added to the pre-trained CNN, works almost as well as CNN trained from scratch on the target turbine. For the cross-wind-farm case, there are more non-linear domain shifts to consider as well. Here, they added both the LR model as well as a CNN model that predicts the error between the input and the output of the LR. Simply fine-tuning the original CNN after pretraining did not yield better results, as it can easily overfit.

Another application of \ac{tl} for anomaly detection in wind turbines is presented by Dimitrov et al. \cite{Schroder.2022}. The authors use \ac{tl} to create a physics-informed machine-learning model, by transferring knowledge from a model trained on simulations to a target model trained on real data. The two models tested are a five-layer dense neural network (ANN) and a five-layer denoising autoencoder (DAE). Both use six SCADA parameters, of which the electrical power is used as a health indicator. The prediction performance of the ANN increases significantly with TL, even when only one month of SCADA data is used for fine-tuning. The power prediction of the DAE did not improve a lot, as it already had very high performance. The anomaly detection ability of the DAE was however not tested.

To enable the use of structurally incomplete data (where only some sensor values are available due to different sample rates), Chen et al. use \ac{tl} to train fault classification models for different types of incomplete data and a model based on only the complete data \cite{Chen.2019}. The information is then transferred between these models to optimally use all available data, instead of only a subset (the structurally complete data). The proposed method has a higher accuracy than the model solely trained on structurally complete data and the resulting models can be applied to all data.

Wen et al. use \ac{tl} to transfer information from univariate time series models to multivariate time series models for anomaly detection \cite{Wen.31.05.2019}. The model is based on an encoder-decoder network (U-Net) and extended for multivariate time series. The model for multivariate time series has encoders for each input feature, to which the information from the models for univariate time series is transferred. The resulting \ac{tl}models all outperformed the models without TL.

Recently, Simani et al. study \ac{tl} for anomaly detection in \ac{scada} data of wind turbines \cite{simani2023}. They focus improving the accuracy of the \ac{nbm}'s output signals and show that this indeed is possible between different wind turbines in the same wind farm. However, the application to fault diagnosis or anomaly detection is not investigated.

\section{Method}\label{method}
In this section, the data, model and transfer learning methods are described.

\subsection{Data}\label{data}
This study utilises data from two offshore wind farms, WF \textit{A} and WF \textit{B}, located in the German North Sea. These wind farms feature different types of turbines. The dataset used in this study is derived from \ac{scada} data and spans several years. It consists of ten-minute-average values from 240 sensors for WF \textit{A} and 30 sensors for WF \textit{B}. The sensors capture various measurements, including power production, wind speed and direction, as well as temperature measurements of bearings.

In addition to the \ac{scada} data, the dataset includes information about the \ac{opmode} for each timestamp. Examples of \acp{opmode} include normal operation, derated operation, service, downtime, and others. For the purposes of this study, only 'normal operation' is considered as normal behaviour, while all other \acp{opmode} are categorised as 'anomalous'.

\subsection{Anomaly detection model}\label{modeling}
For the anomaly detection model, we use a similar setup as presented in \cite{Roelofs.2021, LutzA.etal..}. 

First data representing normal behaviour is selected based on the \ac{opmode}. In addition, only data with power $P > 0$. Due to communication issues the \ac{opmode} is not always properly closed, resulting in some downtimes being missed and logged as, for example, 'normal operation'. Furthermore, we filter out any values where the power is high but the wind speed is outside the cut-in and cut-off values of the wind turbine. The data filtered out is labeled as 'anomalous'.

Next, data is prepared for the \ac{ae} model. First, we remove any features that represent counter values (such as the number of cable windings or the energy production since commissioning of the wind turbine) or set points. Then, we eliminate features that are constant or have up to 3 unique values (indicating low variance). The data does not contain missing values, as they are replaced with zero values. Consequently, we also remove features that have more than 80\% zero values. For features representing angles, we transform them into sine and cosine values to ensure their continuity. Finally, we scale the data between 0 and 1 using min-max-scaling.

The \ac{ae} model used is a fully connected feed forward neural network with a small bottleneck layer and employs the PReLU \cite{he2015delving} activation function. For WF \textit{A} the network consists of 7 layers, where the hidden layers have 200, 100, 50, 100 and 200 units respectively. For WF \textit{B}, there are 3 hidden layers, with 25, 10, and 25 units. 
The network's parameters are optimised using the Adam optimisation algorithm \cite{Kingma.22.12.2014} with a learning rate of 0.001.

To determine whether the model detected an anomaly, we calculate the \ac{rmse} of the reconstruction errors for each timestamp in the training data. This score is referred to as the Anomaly Score $S$. Next, we set a threshold for the training data to maximise the $F_{1/2}$-score. The anomaly labels used for optimisation are based on the \ac{opmode} and power values as described before for normal data selection. We use the $F_{1/2}$-score instead of the $F_1$-score because we aim to prioritise precision over recall for our model.

\subsection{Transfer learning}\label{transfer_learning_methods}
In this paper, we use two types of source models: models trained on 12 months of data of one wind turbine and models trained on 12 months of data of multiple wind turbines. 
These models are then transferred to the target wind turbine using the following three methods:
\begin{itemize}
    \item \textbf{Threshold}: the \ac{ae} of the source model is applied as is. Only the anomaly score threshold is updated.
    \item \textbf{Decoder}: In addition to updating the threshold, the decoder of the \ac{ae} is fine-tuned while keeping the encoder fixed. We use a learning rate of 0.001 and train for 10 epochs.
    \item \textbf{AE}: In this case the complete \ac{ae} network is fine-tuned using a learning rate of 0.0001 (1/10 of original learning rate) and 10 epochs. Afterwards, the threshold is updated.
\end{itemize}

From this point on, we will refer to the two types of TL models as \textit{asset-to-asset} and \textit{multi-asset} models.

\subsection{Evaluation}
The anomaly detection capability of the model is evaluated using the $F_{1/2}$-score on one month of data immediately following the training or tuning data. We determine whether anomalies were correctly detected by using the labels "anomaly" and "normal" based on the \ac{opmode}, wind, and power measurements, as explained in Section \ref{modeling}.

We compare the performance of tuned models with a baseline model trained on 12 months of data.
Furthermore, we conduct three case studies to demonstrate that the transferred models are still effective in detecting anomalous behaviour due to failures. 
Here, we use the criticality $C$ measure as defined in \cite{LutzA.etal..} to evaluate the anomalies detected. The criticality value increases by one if an anomaly is detected during normal operation, decreases by one if no anomaly is detected during normal operation, and remains constant if the \ac{opmode} is not normal. The range of the criticality value is limited to [0, 1000].

\section{Results and discussion}\label{results}
First, baseline models for 10 wind turbines of WF \textit{A} and 6 wind turbines of WF \textit{B} were created to compare the TL models.

\begin{figure}%
    \centering
    \subfloat[\centering WF \textit{A}]{{\includegraphics[width=0.42\textwidth]{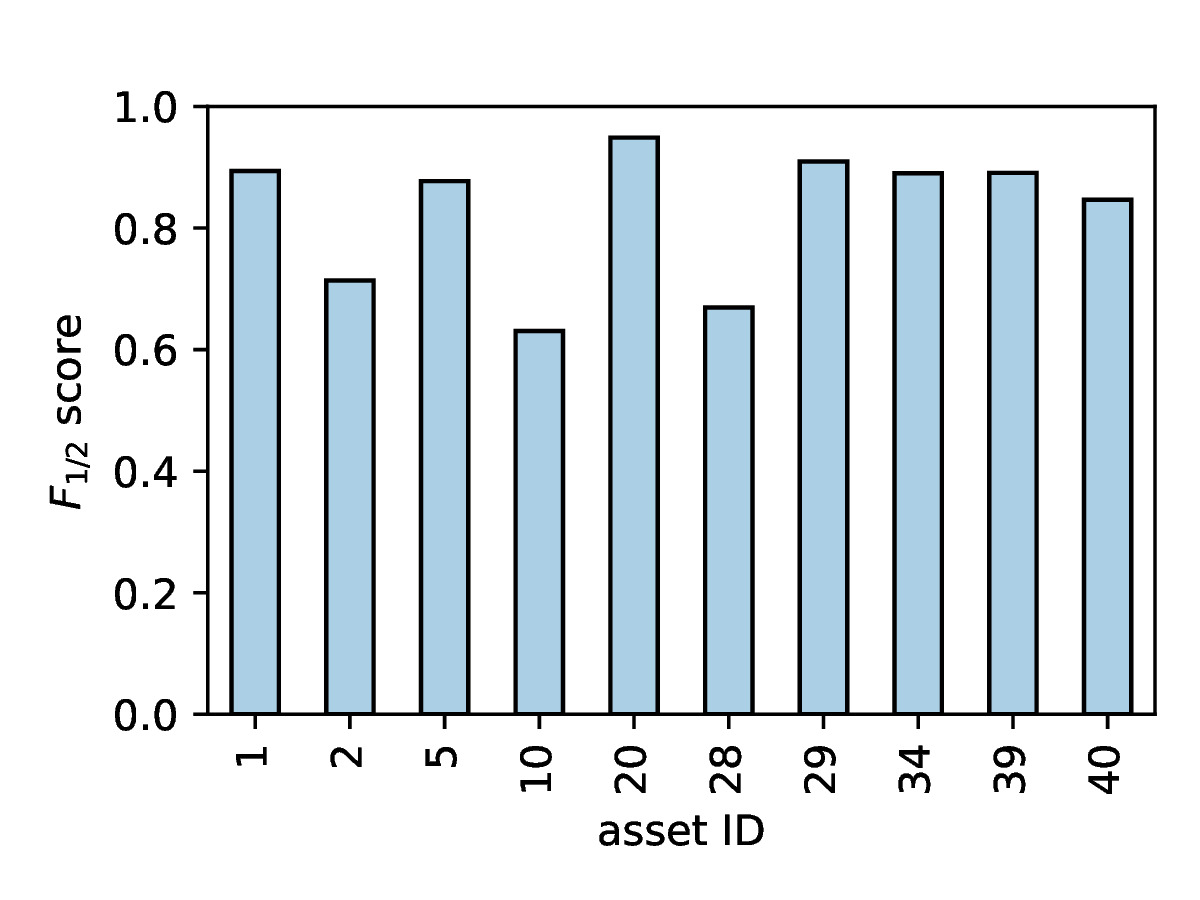}}}%
    \qquad
    \subfloat[\centering WF \textit{B}]{{\includegraphics[width=0.42\textwidth]{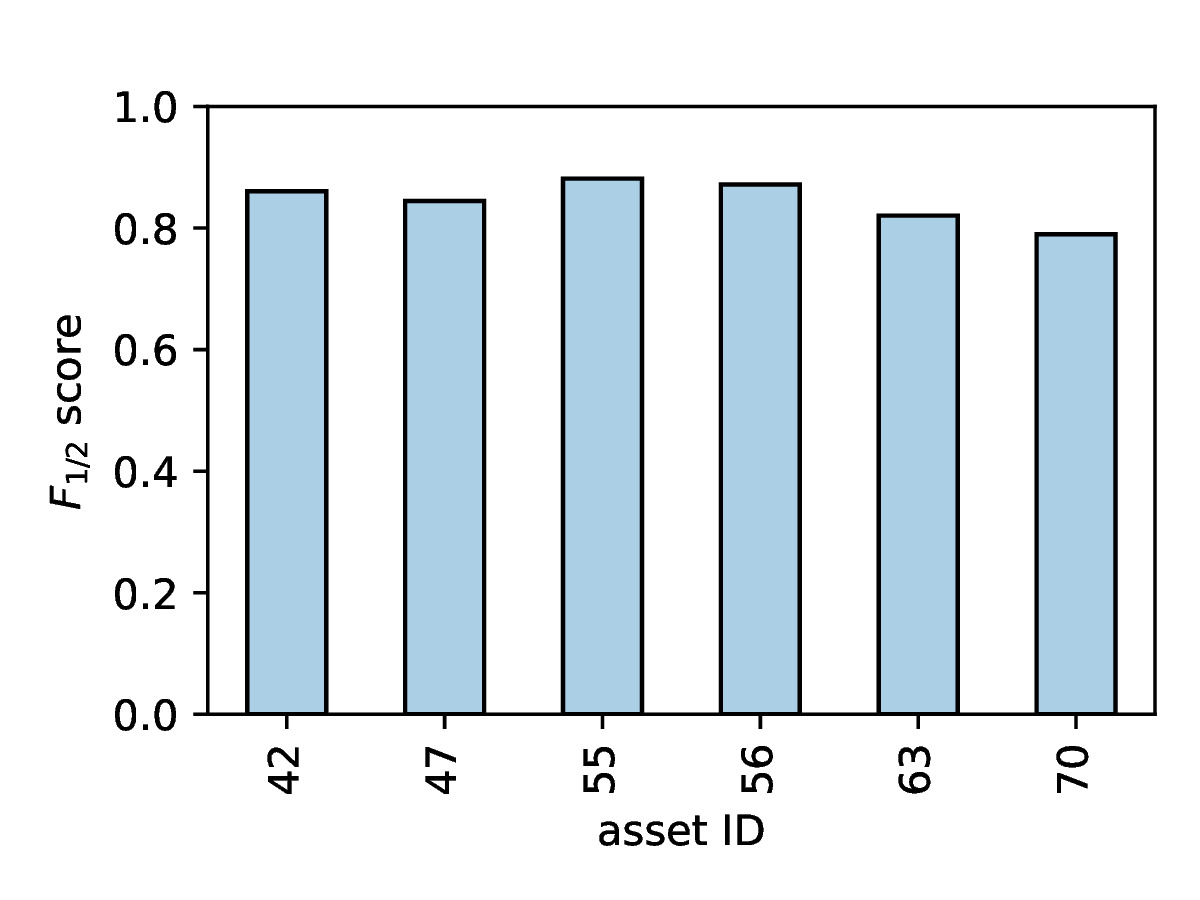} }}%
    \caption{Performance of the baseline models.}%
    \label{fig:baseline}%
\end{figure}

Both the baseline models as well as the source models were trained 1 year of \ac{scada} data representing normal behaviour. The average performance of these baseline models by amount of training data is shown in figure \ref{fig:baseline}. For WF \textit{A} wind turbines 1 and 40 are on opposite sides of the wind farm, whereas 28 and 29 are in the middle. For WF \textit{B} wind turbines 42 and 70 are opposite sides, wind turbines 55 and 56 are in the middle of the wind farm.

Next, the source models were transferred using the methods described in section. \ref{transfer_learning_methods}. In Section \ref{application_1} and Section \ref{application_2} the results for the \textit{asset-to-asset} and \textit{multi-asset} models are described respectively. Finally, in Section \ref{case_studies} the case studies are described and analysed.

\subsection{Asset to asset transfer learning}\label{application_1}
\begin{figure}
    \centering
    \includegraphics[width=\textwidth]{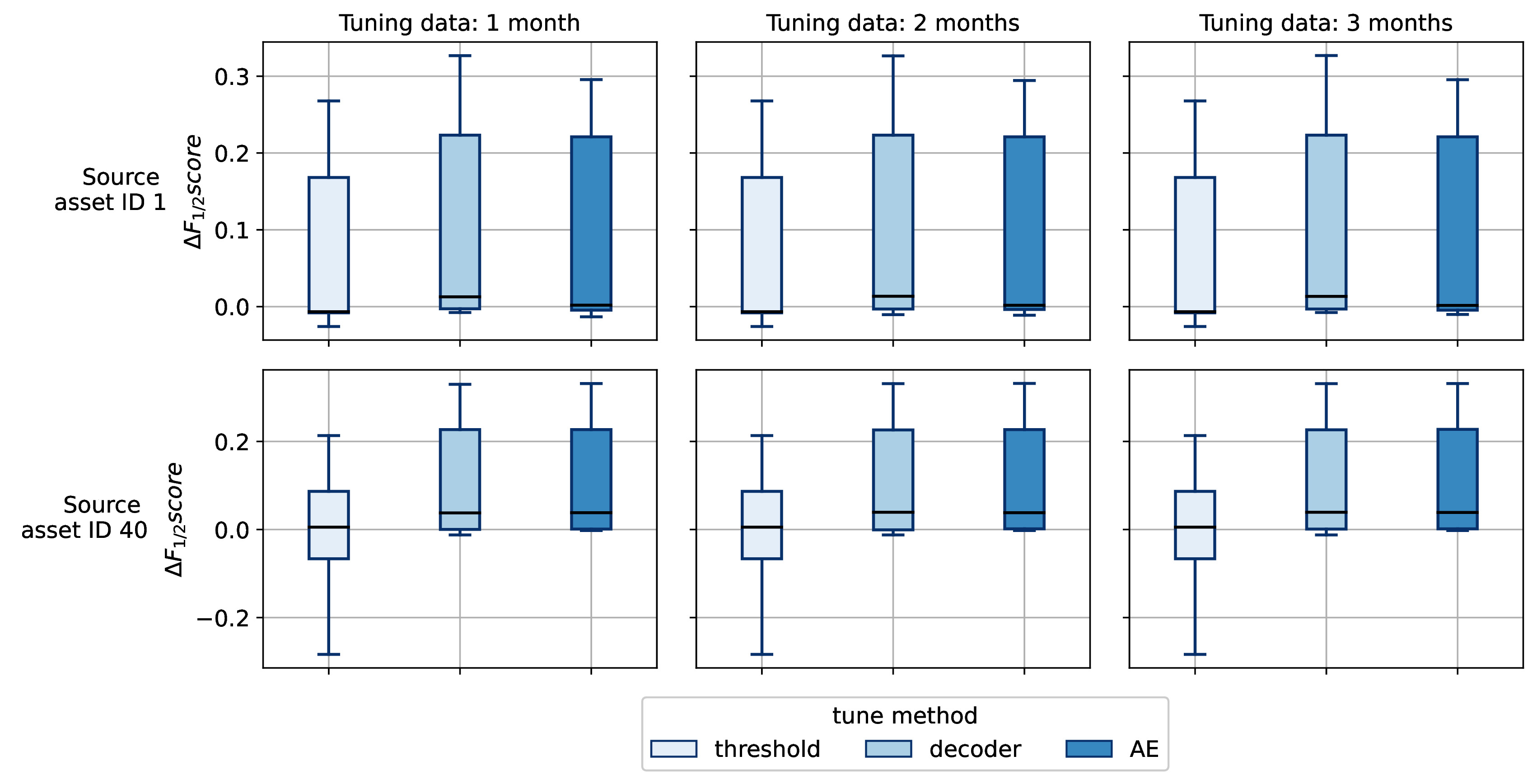}
    \caption{Transfer Learning results for WF \textit{A}. The difference in $F_{1/2}$ score with respect to the baseline model for each wind turbine is shown.}
    \label{fig:asset2assetWFA}
\end{figure}

\begin{figure}
    \centering
    \includegraphics[width=\textwidth]{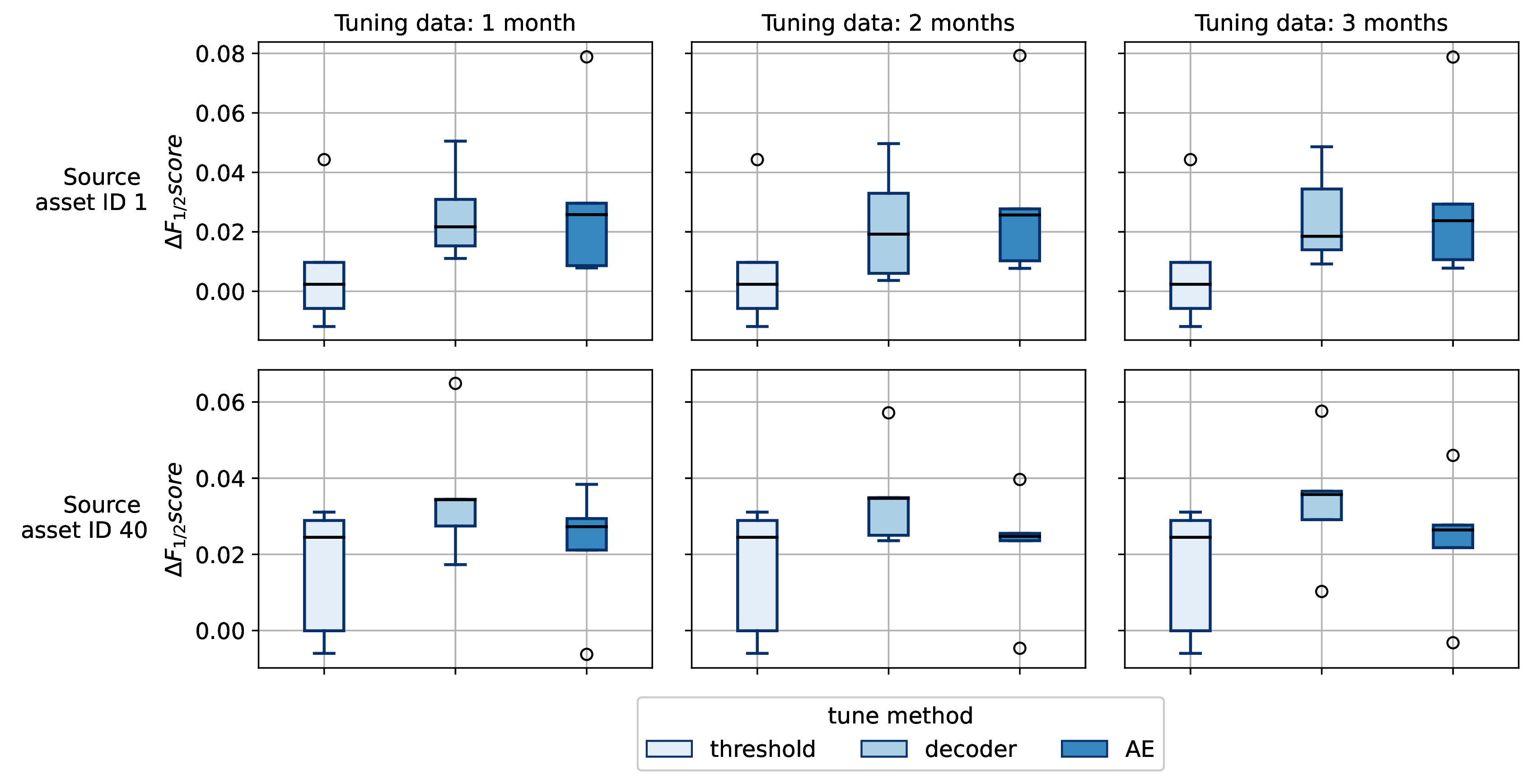}
    \caption{Transfer Learning results for WF \textit{B}. The difference in $F_{1/2}$ score with respect to the baseline model for each wind turbine is shown.}
    \label{fig:asset2assetWFB}
\end{figure}

For WF \textit{A}, the source models were based on data from wind turbines 1 and 40, which are located on opposite sides of the wind farm. The purpose was to investigate whether the location has an influence on the model's performance on the target wind turbines. We will refer to these wind turbines as "source wind turbines" from now on. For WF \textit{B}, the same setup was used, using source wind turbines 42 and 70.

Results are shown in Figure \ref{fig:asset2assetWFA} and Figure \ref{fig:asset2assetWFB} for TL models based on only a small amount of tuning data, ranging from 1 to 3 months. Notably, these results show that it is indeed possible to create an anomaly detection model with a relatively small amount of data. The F$_{1/2}$ scores of the TL models are comparable or slightly better than the baseline models, which were trained on a full year of \ac{scada} data. Whether 1 month or 3 months of tuning data is used, does not seem to have a significant effect.

Interestingly, the location of the chosen source wind turbines does not appear to have a significant impact on the model's performance. The TL models trained using data from wind turbine 1 fit the data of wind turbine 40, on the opposite side of the wind farm, quite well, even with just one month of tuning data. The same is observed for WF \textit{B}. There is however not enough data to state that one can simply use any of the wind turbines as source, since the data quality of the source wind turbine overshadows this effect. Therefore, we use one of the middle assets in the WFs with a good baseline model performance as source wind turbine for the case studies.

The transferred models show promising results, performing on par or slightly better than the baseline models. It is worth noting that the autoencoder of the source model is capable of reconstructing the data of the target wind turbines, even without tuning the autoencoders. These models can yield good model scores, even though the autoencoder was trained on data from a different wind turbine, showing that the wind turbines within the same wind farm behave very similarly, since the normal behaviour primarily depends on the environment.

In some cases, the transferred models even outperform the baseline models. The improved performance of the transferred models may be attributed to their ability to fill knowledge gaps in the baseline models, particularly when the training data of the baseline models lacks normal behaviour within a period (e.g. winter or summer) similar to the prediction period.

Fine-tuning the decoder or the complete autoencoder further enhances the models' performance in terms of F$_{1/2}$ score compared to the threshold method, as depicted in Figures \ref{fig:asset2assetWFA} and \ref{fig:asset2assetWFB}. Specifically, fine-tuning the decoder alone outperforms the other TL methods in most cases. However, we find that caution must be exercised when tuning the autoencoder, as it could lead to overfitting and negative transfer.

\subsection{Multi-asset model transfer learning}\label{application_2}
\begin{figure}%
    \centering
    \subfloat[\centering WF \textit{A}]{{\includegraphics[width=\textwidth]{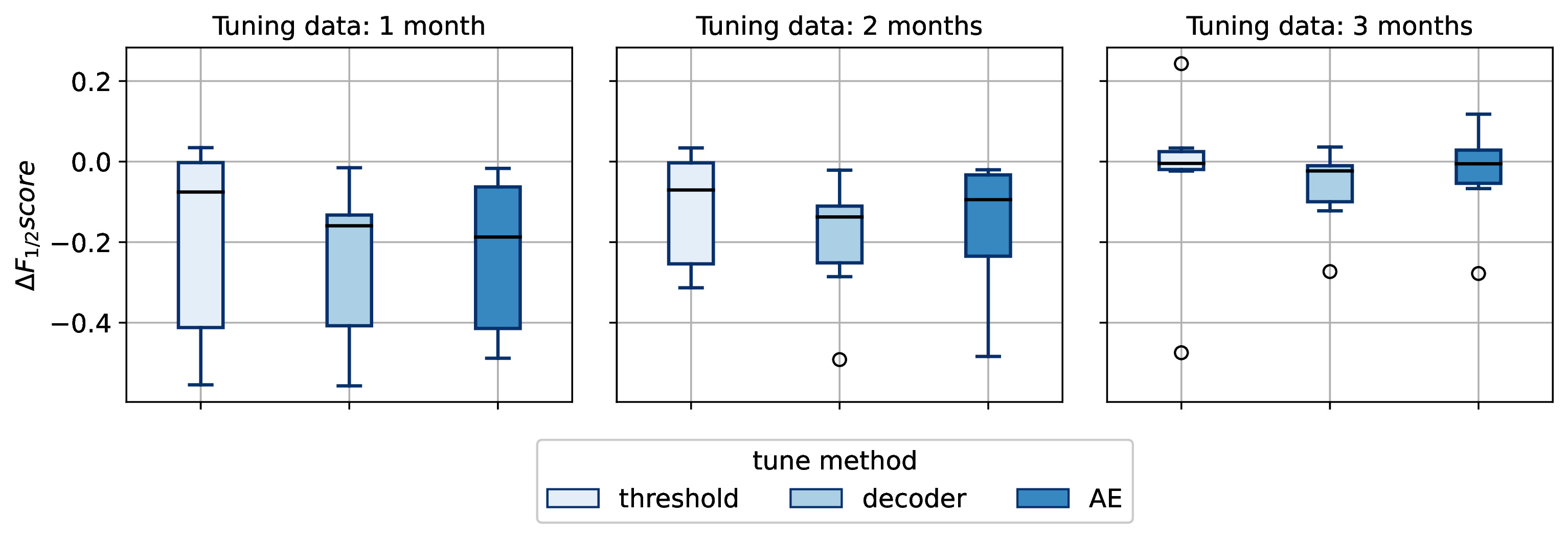}}}%
    
    \subfloat[\centering WF \textit{B}]{{\includegraphics[width=\textwidth]{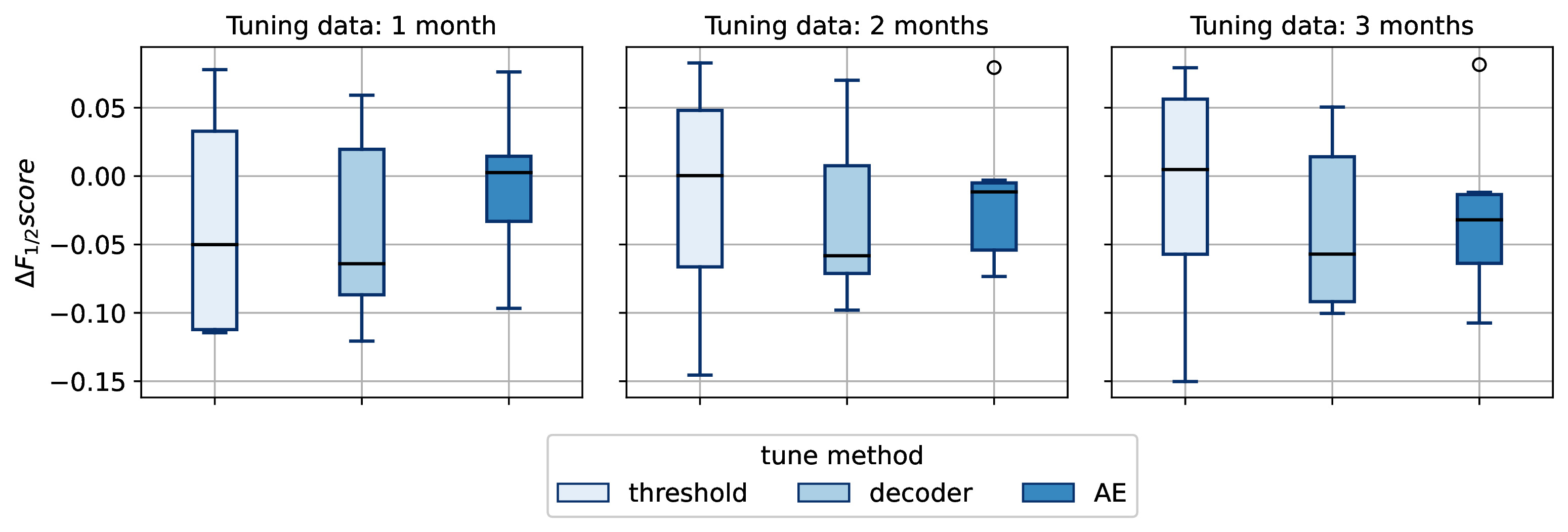} }}%
    \caption{Transfer Learning results using data of multiple assets to train a source model and fine-tuning on the target wind turbine. The difference in $F_{1/2}$ score with respect to the baseline model for each wind turbine is shown.}%
    \label{fig:multiTL}%
\end{figure}

In this section, we discuss the results for the \textit{multi-asset} TL models. Here, the source models were trained on 9 wind turbines of W$_A$ and fine-tuned on the target wind turbine, which is not part of the source model's training data. For WF \textit{B} the source models were trained on data from 5 wind turbines and fine-tuned on the target wind turbine. Results are shown in Figure \ref{fig:multiTL}.

Compared to the results of the \textit{asset-to-asset} models, these models do not perform as well, particularly for WF \textit{A}. 
Since the wind turbines within one wind farm behave very similar, this could be caused by overfitting, as the model sees almost the same data multiple times.
In case of WF \textit{A}, the TL models for all wind turbines except 40 perform significantly better with 3 months of tuning data. For WF \textit{B} the models' performance does not seem to be significantly influenced by the amount of tuning data.

\subsection{Case Studies}\label{case_studies}
For the case studies we use the same set up as before, where a baseline model is trained on one year of data of the target wind turbine to compare with TL models. The source models are trained with data of a wind turbine in the middle of the wind farm, for which the baseline model performed well. For WF \textit{A} this is wind turbine 29, for WF \textit{B} wind turbine 55. Next, the models' decoders were fine-tuned with 1 to 2 months of data of the target wind turbines.

The first two case studies are for wind turbines of WF \textit{A}, the third case study contains a failure in a wind turbine in WF \textit{B}.

\subsubsection{Case study 1}
\begin{figure}
    \centering
    \includegraphics[width=0.8\textwidth]{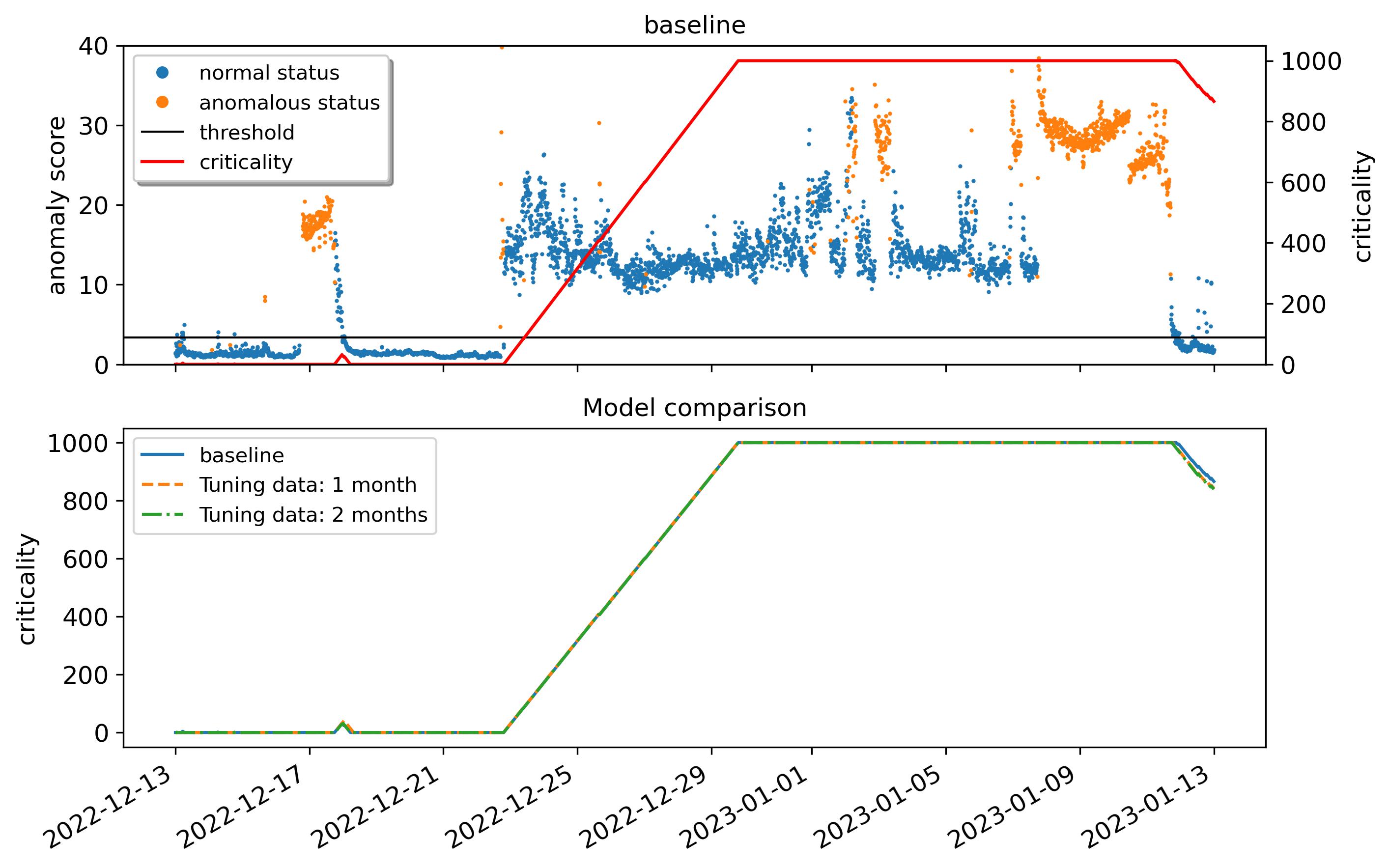}
    \caption{Case Study 1: Rotor brake and hydraulic problems for a wind turbine of WF \textit{A} cause a change point in some sensor measurements. The anomalous behaviour is detected correctly by TL models using only 1 to 2 months of data for fine-tuning. The upper graph depicts the anomaly score, threshold and resulting criticality measure for the baseline model, the lower graph compares the criticality for the three models.}
    \label{fig:rotorbrake}
\end{figure}

Anomalies are detected from 2022-12-22 onward, which were caused by problems in the rotor brake and hydraulic system. The anomalies are caused by several related sensor measurements suddenly behaving very differently. Both TL models are able to detect this anomaly easily, as seen in Figure \ref{fig:rotorbrake}. The anomaly score during abnormal operation on 2022-12-27 is lower for the TL models compared the anomaly score of the baseline model, indicating that the TL models are less sensitive than the baseline model. This is however expected, since only a small amount of target data was used for tuning.

\subsubsection{Case study 2}
\begin{figure}
    \centering
    \includegraphics[width=0.8\textwidth]{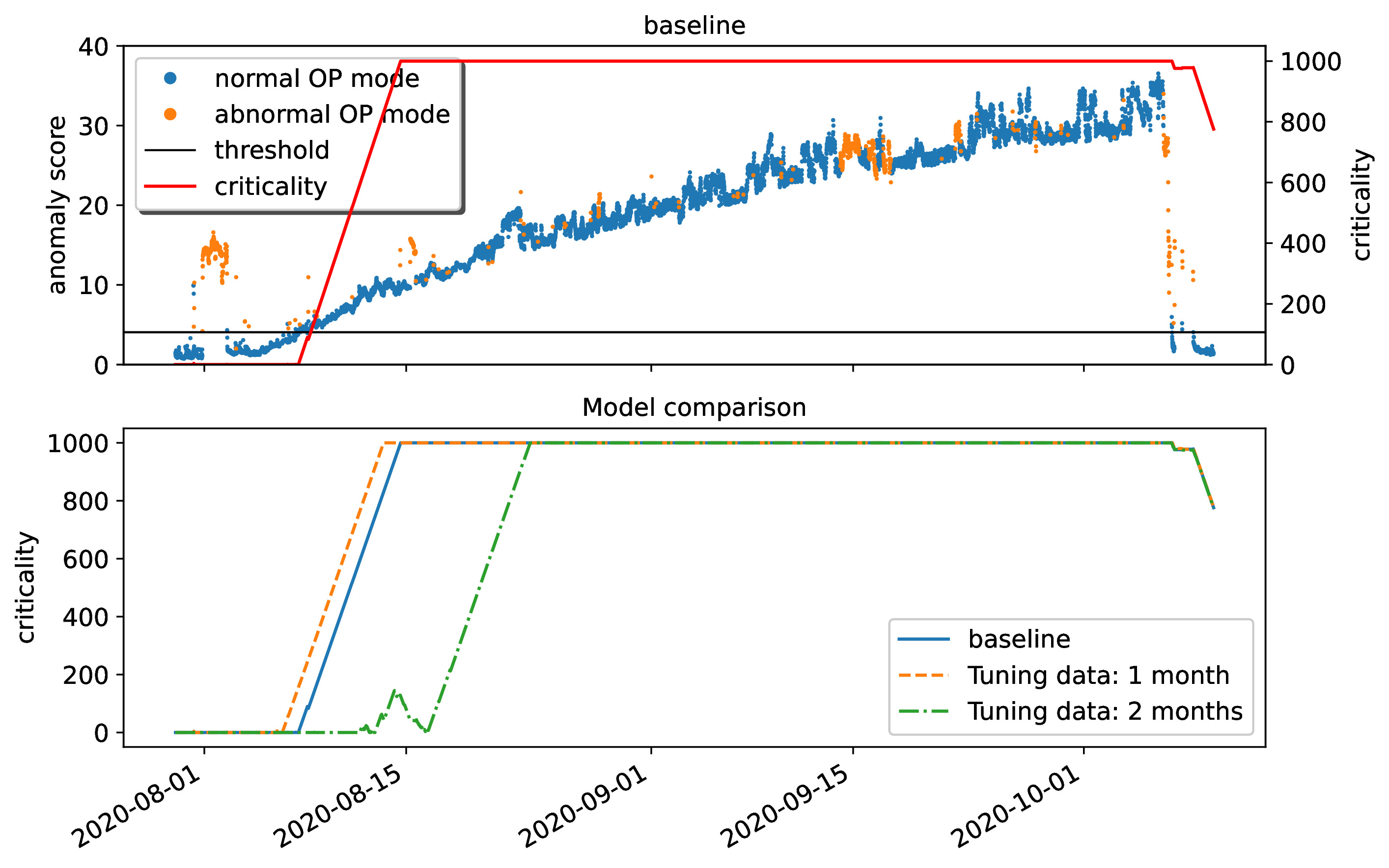}
    \caption{Case Study 2: Sensor drift caused by problems in the water cooling system of a wind turbine of WF \textit{A} cause a change point in some sensor measurements. The anomalous behaviour is detected correctly by TL models using only 1 to 2 months of data for fine-tuning. The upper graph depicts the anomaly score, threshold and resulting criticality measure for the baseline model, the lower graph compares the criticality for the three models.}
    \label{fig:water_cooling}
\end{figure}

Problems in the converter cooling system causes a sensor drift, which in turn causes a positive trend in the anomaly score of the baseline model, s. Figure \ref{fig:water_cooling}. After maintenance on 2022-10-07 the problem was fixed and the anomaly score normalises. Both tuned models are able to detect the problem, although they detect the anomaly later than the baseline model. Interestingly, the model fine-tuned with 2 months of data is less sensitive to this behaviour and detects the anomaly a week later. Nevertheless, the anomaly could be detected weeks before maintenance takes place. Also, the anomaly score shows a positive trend, before the threshold is exceeded.

\subsubsection{Case study 3}
\begin{figure}
    \centering
    \includegraphics[width=0.8\textwidth]{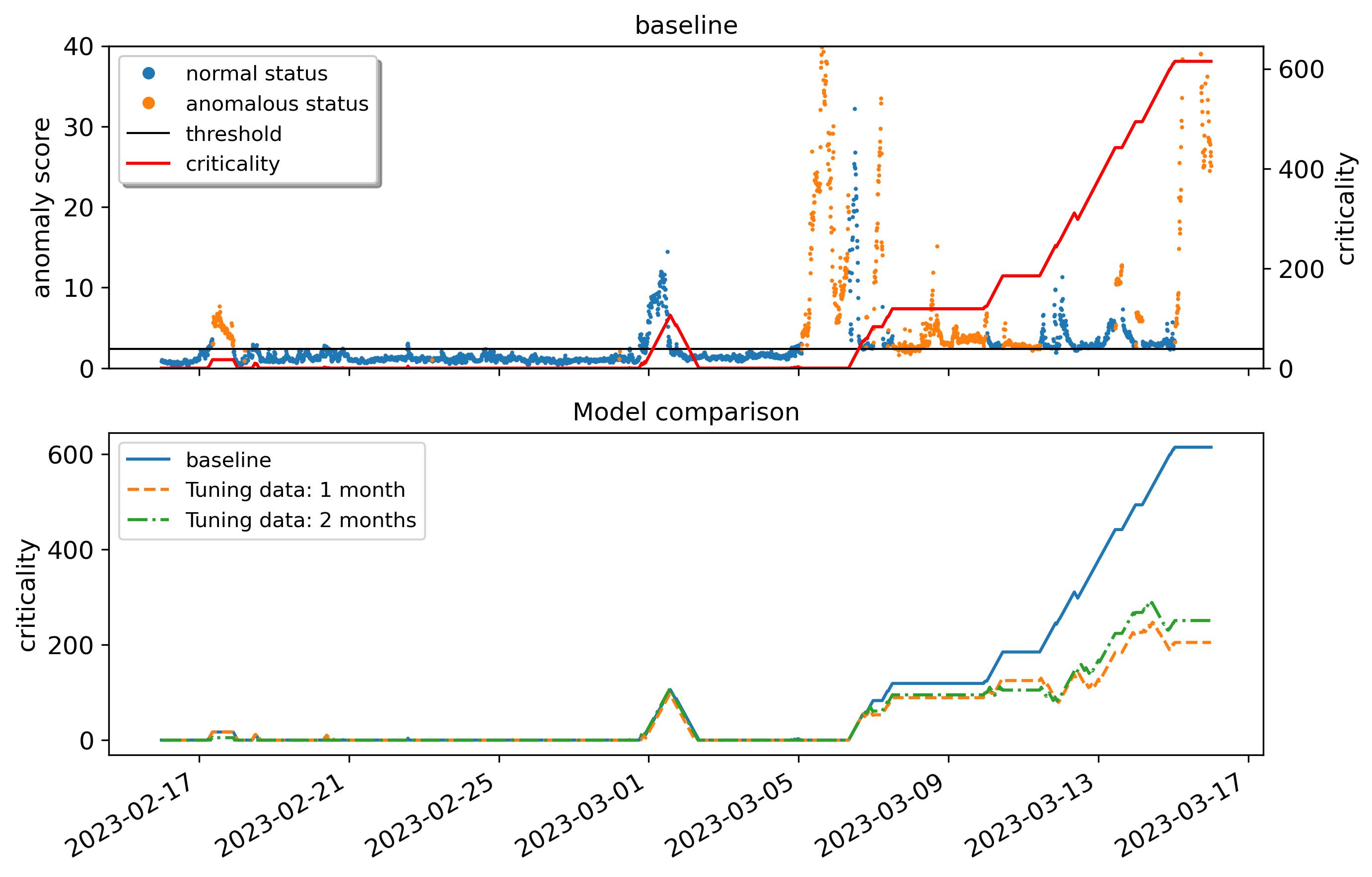}
    \caption{Case Study 3: Anomalies cause by a failure in the gearbox of a wind turbine of WF \textit{B}. Using only 1 to 2 months of data for fine-tuning, the anomaly detection model is capable of detecting the failure correctly. The upper graph depicts the anomaly score, threshold and resulting criticality measure for the baseline model, the lower graph compares the criticality for the three models.}
    \label{fig:gearbox}
\end{figure}

Anomalies are detected between 2023-02-28 and 2023-03-27 for a wind turbine in WF \textit{B}, as shown in Figure \ref{fig:gearbox}. The cause of the anomalies was a failure in the gearbox of the wind turbine. Both TL model are able to detect some anomalous behaviour on 2023-03-01, after which the anomaly score normalises for a short time. Next, the wind turbine is regularly in derated operation. Afterwards the turbine was down for maintenance and the gearbox was replaced. 

\section{Conclusions}\label{summary}
To investigate the feasibility of autoencoder-based anomaly detection with limited training data for wind turbines, two transfer learning applications using tree methods were conducted. The study focused on anomaly detection within a single wind farm, using different amounts of tuning data and wind turbine types from two wind farms.

Two types of TL models were studied, \textit{asset-to-asset} and \textit{multi-asset} TL models. In case of \textit{asset-to-asset} the source models were trained on a single wind turbine and fine-tuned on data of the target wind turbine. For the \textit{multi-asset} TL models, the source models were trained on data of multiple wind turbines within the same wind farm and fine-tuned on data of the target wind turbine.

In both cases, the TL models fine-tuned using 1 to 3 months of data performed as well or slightly better than the baseline models, which were trained on a full year of data. The \textit{asset-to-asset} TL models showed superior performance compared to the \textit{multi-asset} TL models. Additionally, three TL methods were investigated: only updating the threshold, fine-tuning the decoder and fine-tuning the autoencoder, of which the decoder-method showed the best results.

To demonstrate that early fault detection in wind turbines is indeed possible even with a limited amount of training data three case studies were conducted. In all three cases the failure was detected by TL models tuned on only 1 to 2 months of data, highlighting the potential of TL for improving anomaly detection in wind turbines, reducing the amount of data and resources needed for effective detection. 

\paragraph{Acknowledgements}
The development of methods presented was funded by the German Federal Ministry for Economic Affairs and Climate Action (BMWK) through the research project “ADWENTURE” (FKZ 03EE2030).



\bibliographystyle{elsarticle-num} 
\bibliography{references}

\end{document}